# Anomaly Detection in High Dimensional Data

**Priyanga Dilini Talagala**
Department of Econometrics and Business Statistics, Monash University, Australia, and
ARC Centre of Excellence for Mathematics and Statistical Frontiers
Email: dilini.talagala@monash.edu
Corresponding author

**Rob J. Hyndman**
Department of Econometrics and Business Statistics, Monash University, Australia, and
ARC Centre of Excellence for Mathematics and Statistical Frontiers

**Kate Smith-Miles**
School of Mathematics and Statistics, University of Melbourne, Australia, and
ARC Centre of Excellence for Mathematics and Statistical Frontiers

# Anomaly Detection in High Dimensional Data

**Abstract**

The HDoutliers algorithm is a powerful unsupervised algorithm for detecting anomalies in high-dimensional data, with a strong theoretical foundation. However, it suffers from some limitations that significantly hinder its performance level, under certain circumstances. In this article, we propose an algorithm that addresses these limitations. We define an anomaly as an observation that deviates markedly from the majority with a large distance gap. An approach based on extreme value theory is used for the anomalous threshold calculation. Using various synthetic and real datasets, we demonstrate the wide applicability and usefulness of our algorithm, which we call the `stray` algorithm. We also demonstrate how this algorithm can assist in detecting anomalies present in other data structures using feature engineering. We show the situations where the stray algorithm outperforms the HDoutliers algorithm both in accuracy and computational time. This framework is implemented in the open source R package `stray`.

## 1 Introduction

The problem of anomaly detection has many different facets, and detection techniques can be highly influenced by the way we define anomalies, type of input data and expected output. These differences lead to wide variations in problem formulations, which need to be addressed through different analytical techniques. Although several useful computational methods currently exist, developing new methods for anomaly detection continues to be an active, attractive interdisciplinary research area owing to different analytical challenges in various application fields, such as environmental monitoring (Talagala et al. 2019b; Leigh et al. 2019), object tracking (Gupta et al. 2014; Sundaram et al. 2009), epidemiological outbreaks (Gupta et al. 2014), network security (Hyndman, Wang & Laptev 2015; Cao et al. 2015) and fraud detection (Talagala et al. 2019a). Ever-increasing computing resources and advanced data collection technologies that emphasise real-time, large-scale data are other reasons for this growth since they introduce new analytical challenges with their increasing size, speed and complexity that demand effective, efficient analytical and computing techniques.

Anomaly detection has two main objectives, which are conflicting in nature: One downgrades the value of anomalies and attempts eliminating them, while the other demands special attention be paid to anomalies and root-cause analysis be conducted. The presence of anomalies in data can be considered data flaws or measurement errors that can lead to biased parameter estimation, model misspecification and misleading results if classical analysis techniques are blindly applied (Ben-Gal 2005; Abuzaid, Hussin & Mohamed 2013). In such situations, the focus is to find opportunities to remove anomalous points and thereby improve both the quality of the data and results from the subsequent data analysis (Novotny & Hauser 2006). In contrast, in many other applications, anomalies themselves are the main carriers of significant and often critical information, such as extreme weather conditions (e.g., bushfire, tsunami, flood, earthquake, volcanic eruption and water contamination), faults and malfunctions (e.g., flight tracking and power cable tracking) and fraud activities (Ben-Gal 2005), that can cause significant harm to valuable lives and assets if not detected and treated quickly.

High-dimensional datasets exist across numerous fields of study (Liu et al. 2016). Some anomaly detection algorithms also use feature engineering as a dimension reduction technique and thereby convert other data structures, such as a collection of time series using time series features (Talagala et al. 2019b; Hyndman, Wang & Laptev 2015), collection of scatterplots using scagnostics (Wilkinson, Anand & Grossman 2005) and genomic micro arrays and chemical compositions in biology (Liu et al. 2016) into high-dimensional data prior to the detection process for easy control. Under the high-dimensional data scenario, all attributes can be of the same data type or a mixture of different data types, such as categorical or numerical, which has a direct impact on the implementation and scope of the algorithm. Much research attention has been paid to anomaly detection for numerical data (Breunig et al. 2000; Tang et al. 2002; Jin et al. 2006; Gao et al. 2011). Limited methods are available that treat both numerical and categorical data using correspondence analysis, for example, as in Wilkinson (2017).

High-dimensional anomalies can arise in all the attributes or a subset of the attributes (Unwin 2019). If all anomalies in a high-dimensional data space were anomalies in a lower dimension, then anomaly detection can be performed using axis parallel views or by incorporating an additional step of variable selection for the detection process. However, in practice, certain high-dimensional instances are only perceptible as anomalies if treated as high-dimensional problems and the correlation structure of all the attributes considered. Otherwise, these tend to be overlooked if attributes are considered separately (Wilkinson 2017; Ben-Gal 2005).

The problem of anomaly detection has been extensively studied over the past decades in many application domains. Several surveys of anomaly detection techniques have been conducted in general (Chandola, Banerjee & Kumar 2009; Aggarwal 2017) or for specific data domains such as high-dimensional data, network data (Shahid, Naqvi & Qaisar 2015), temporal data (Gupta et al. 2014), machine learning and statistical domains (Hodge & Austin 2004), novelty detection (Pimentel et al. 2014), intrusion detection (Sabahi & Movaghar 2008) and uncertain data (Aggarwal & Yu 2008). Some algorithms are application specific and take advantage of the underlying data structure or other domain-specific knowledge (Talagala et al. 2019b). More general algorithms without domain-specific knowledge are also available with their own strengths and limitations (Breunig et al. 2000; Tang et al. 2002; Jin et al. 2006; Gao et al. 2011). Among the many possibilities, the HDoutliers algorithm, recently proposed by Wilkinson (2017), is a powerful unsupervised algorithm, with a strong theoretical foundation, for detecting anomalies in high-dimensional data. Although this algorithm has many advantages, a few characteristics hinder its performance. In particular, under certain circumstances it tends to increase the rate of false negatives (i.e., the detector ignores points that appear to be real anomalies) because it uses only the nearest-neighbour distances to distinguish anomalies. Further, to deal with large datasets with numerous observations it uses the Leader algorithm (Hartigan & Hartigan 1975), which forms several clusters of points in one pass through the dataset using a ball of a fixed radius. By incorporating this clustering method, it tries to gain the ability to identify anomalous clusters of points. However, in the presence of very close neighbouring anomalous clusters it tends to increase the rate of false negatives. Further, this additional step of clustering has a serious negative impact on the computational efficiency of the algorithm when dealing with large datasets.

Through this study, we make three fundamental contributions. First, we propose an algorithm called *stray*, representing '**S**earch and **TR**ace **A**nomal**Y**', that addresses the limitations of the HDoutliers algorithm. The stray algorithm presented here focuses specifically on fast, accurate anomalous score calculation using simple but effective techniques for improved performance. Second, we introduce an R (R Core Team 2019) package, `stray` (Talagala, Hyndman & Smith-Miles 2019), that implements the stray algorithm and related functions. Third, we demonstrate the wide applicability and usefulness of our stray algorithm, using various datasets.

Our improved algorithm, stray, has many advantages: (1) It can be applied to both one-dimensional and high-dimensional data. (2) It is unsupervised in nature and therefore does not require training datasets for the model-building process. (3) The anomalous threshold is a data-driven threshold and has a valid probabilistic interpretation because it is based on the ex- treme value theory. (4) By using k-nearest neighbour distances for anomalous score calculation, it gains the ability to deal with the masking problem. (5) It can provide near real-time support to datasets that stream in large quantities owing to its use of fast nearest neighbour searching mechanisms. (6) It can deal with data that may have multimodal distributions for typical data instances. (7) It produces both score (to indicate how anomalous the instances are) and binary classification (to reduce the searching space during the visual and root-cause analysis) for each data instance as an output. (8) It can detect outliers as well as inliers.

The remainder of this paper is organised as follows. Section 2 presents the related work to lay the foundation for the stray algorithm. Section 3 describes the limitations of the HDoutliers algorithm that hinder its performance. Section 4 presents the improved algorithm, stray, that addresses the limitations of the HDoutliers algorithm. Section 5 presents a comprehensive evaluation, illustrating the key features of the stray algorithm. Section 6 includes an application of stray algorithm related to pedestrian behaviour in the city of Melbourne, Australia. Section 7 concludes the article and presents future research directions.

## 2 Background

### 2.1 Types of Anomalies in High Dimensional Data

The problems of anomaly detection in high-dimensional data are threefold (Figure 1), involving detection of: (a) global anomalies, (b) local anomalies and (c) micro clusters or clusters of anomalies (Goldstein & Uchida 2016). Most of the existing anomaly detection methods for high-dimensional data can easily recognise global anomalies since they are very different from the dense area with respect to their attributes. In contrast, a local anomaly is only an anomaly when it is distinct from, and compared with, its local neighbourhood. Madsen (2018) introduces a set of algorithms based on a density or distance definition of an anomaly, which mainly focuses on local anomalies in high-dimensional data. Micro clusters or clusters of anomalies may cause masking problems. Very little attention has been paid to this problem relative to the other two categories. The recently proposed HDoutliers algorithm (Wilkinson 2017) addresses this problem to some extent by grouping instances together that are very close in the high-dimensional space and then selecting a representative member from each cluster before calculating nearest neighbour distances for the selected instances. In this study, we focus on all three of these anomaly types.

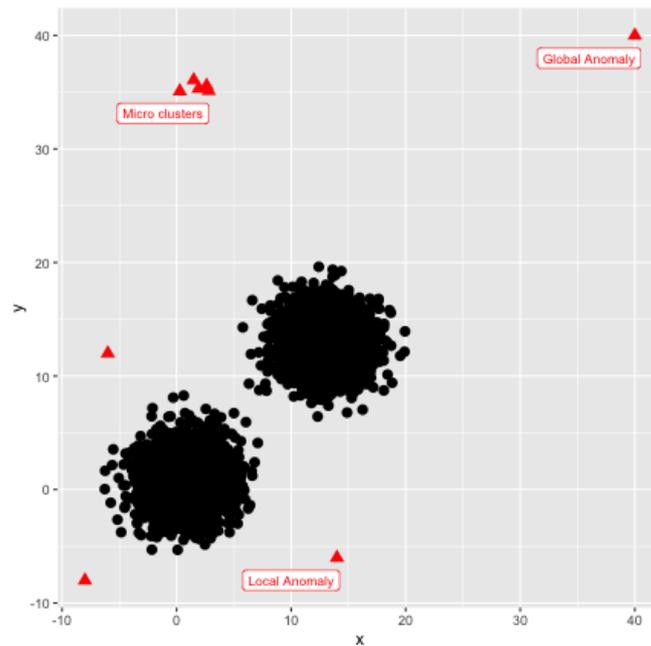

**Figure 1:** *Different types of anomalies in high-dimensional data. Anomalies are represented by red triangles and black dots correspond to the typical behaviour.*

## 2.2 Definitions for Anomalies in High Dimensional Data

Anomalies are often mentioned in the literature under several alternative terms, such as outliers, novelty, faults, deviants, discordant observations, extreme values/cases, change points, rare events, intrusions, misuses, exceptions, aberrations, surprises, peculiarities, odd values and contaminants, in different application domains (Chandola, Banerjee & Kumar 2009; Gupta et al. 2014; Zhang, Wu & Yu 2010). Of these, the two terms anomalies and outliers are used commonly and interchangeably in the literature describing research related to the topic. The term inlier also relates to the topic, but rarely appears in the literature on anomaly detection. Inliers are those points that appear between typical clusters without attaching to any of the clusters, but still lie within the range defined by the typical clusters (Jouan-Rimbaud et al. 1999). In contrast, the corresponding notion of an 'outlier' is generally used to refer to a data instance that appears out of the space more towards the tail of a distribution, defined by the typical data instances. Some classical methods related to the topic fail to detect inliers and only focus on outliers (Jouan-Rimbaud et al. 1999). However, detecting inliers is equally important because they can give rise to interpolation errors. In this study, we focus on both inliers and outliers. To avoid any confusion, we use the term 'anomaly' for the purpose of nomenclature throughout this paper.

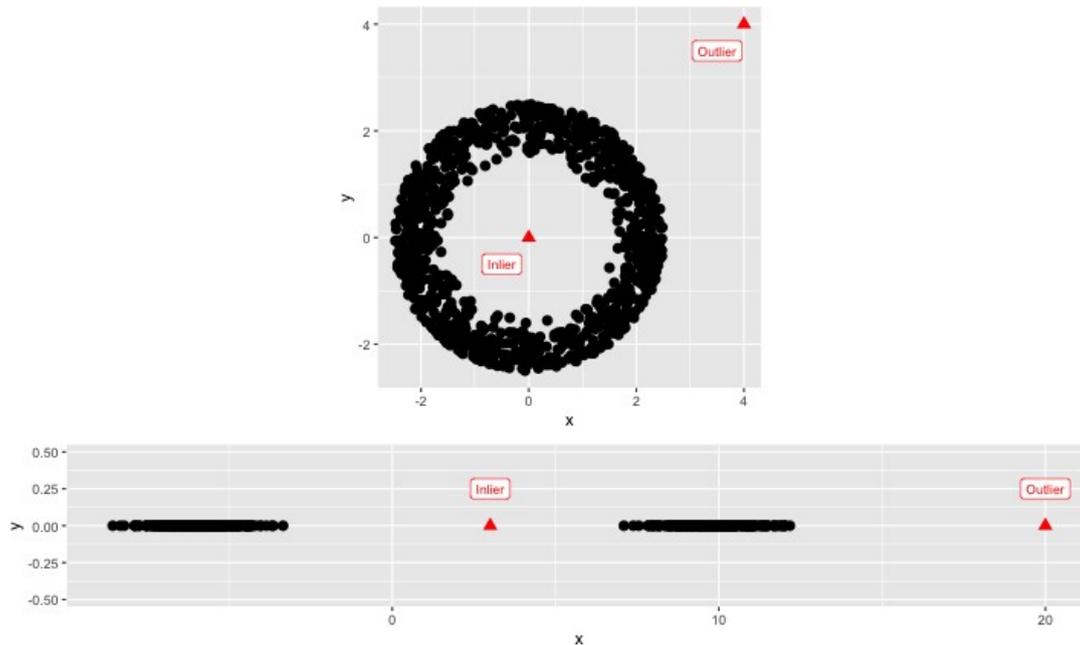

**Figure 2:** *Inliers vs outliers. Anomalies are represented by red triangles and black dots correspond to the typical behaviour.*

Owing to the complex nature of the problem, it is difficult to find a unified definition for an anomaly and the definition often depends on the focus of the study and the structure of the input data available to the system (Williams 2016; Unwin 2019). However, there are some definitions that are general enough to cope with datasets with various application domains. Grubbs (1969) defines an anomaly as an observation that deviates markedly from other members of the dataset. However, this deviation can be defined in terms of either distance or density. Burridge & Taylor (2006), Wilkinson (2017) and Schwarz (2008) have all proposed methods for anomaly detection by defining an anomaly in terms of distance. In contrast, Hyndman (1996), Clifton, Hugueny & Tarassenko (2011) and Talagala et al. (2019a) have proposed methods that define an anomaly with respect to either the density or the chance of the occurrence of observations. Madsen (2018) also provides a series of distance and density-based anomaly detection algorithms.

In this study, we define an anomaly as an observation that deviates markedly from the majority with a large distance gap under the assumption that there is a large distance between typical data and the anomalies compared with the distance between typical data.

# 3 Limitations of HDoutliers Algorithm

Although the HDoutliers algorithm (Wilkinson 2017) has many advantages, a few characteristics limit its possibilities. Next, we discuss these limitations in detail.

## 3.1 HDoutliers Uses Only the Nearest Neighbour Distance to Discriminate Anomalies

The HDoutliers algorithm uses the Leader algorithm (Hartigan & Hartigan 1975) to form small clusters of points, prior to calculating nearest neighbour distance. In the Leader algorithm, each cluster is a ball in the high-dimensional data space. In the HDoutliers algorithm, the radius of this ball is selected such that it is well below the expected value of the distances between $n(n-1)/2$ pairs of points distributed randomly in a $d$-dimensional unit hypercube.

After forming clusters using the Leader algorithm, the HDoutliers algorithm selects representative members from each cluster. It then calculates the nearest neighbour distances for each of these representative members. These distances are then used to identify the anomalies based on the assumption that anomalies bring large distance separations between typical data and the anomalies, in comparison to the separations between typical data themselves. Therefore, under this assumption it is believed that any anomalous cluster will appear far away from the clusters of the typical data points. As a result, the nearest neighbour distance for this anomalous cluster will be significantly higher than that of the clusters of typical data and thereby identify it as an anomalous cluster. All the data points contained in the anomalous cluster are then marked as anomalous points within a given dataset.

However, one further assumption for this method to work properly is that any anomalous clusters present in the dataset are isolated. For example, imagine a situation in which two anomalous clusters are very close to one another but are far away from the rest of the typical clusters. Now, the two clusters will become nearest neighbours to one another and they will jointly project them by being anomalous by giving very small nearest neighbour distances for both clusters that are compatible with the nearest neighbour distances of the rest of the typical clusters. Figures 6 (c-II) and (d-II) further elaborate this argument. In these two examples, the HDoutliers algorithm (with the clustering step) declares points as anomalies only if they are isolated and fails to detect anomalous clusters that share a few cluster neighbours. Although the HDoutliers algorithm incorporates the clustering step with the aim of identifying anomalous clusters of points, because of the very small size of the ball that is used to produce clusters (exemplars) in the *d*-dimensional space, it fails to bring all the points into a single cluster and instead produces a few anomalous clusters that are very close to one another. These anomalous clusters then become nearest neighbours to one another and have very small nearest neighbour distances for the representative member of each cluster. Since the detection of anomalies entirely depends on these nearest neighbour distances and since the anomalous clusters do not show any significant deviation from typical clusters with respect to the nearest neighbour distances, the algorithm now fails to detect these points as anomalies and thereby increases the rate of false negatives.

### 3.2 Problems Due to Clustering Via Leader Algorithm

After forming clusters of data points, the HDoutliers algorithm completely ignores the density of the data points. Once it forms clusters of data points using the Leader algorithm, it selects a representative member from each cluster and carries out further analysis only using these representative members. Figure 6 (e-II) provides an example related to this issue. This dataset is a bimodal dataset with an anomalous point located between the two typical classes. The entire dataset contains 2,001 data points. The data points gathered at the leftmost upper corner represent one typical class with 1,000 data points. The second typical class of data points is gathered at the rightmost bottom corner with another 1,000 data points. Since this second class of data points is closely compacted in substance, the 1,000 data points are now wrapped by a single ball when forming clusters using the Leader algorithm. In the HDoutliers algorithm, the next step is to select one member from each of these clusters. Once it selects a representative member from this ball that contains 1,000 data points, it ignores the remaining 999 data points in detecting anomalies. This step misleads the algorithm, and the remaining steps of the algorithm view this representative member as an isolated data point, although it is surrounded by 999 neighbouring data points in the original dataset. Therefore, all data points in this entire class are declared as anomalies by the algorithm, although it contains half of the dataset. Unwin (2019) suggests jittering not as a perfect solution, but as an alternative to mitigate this problem. Unwin (2019) also argues that the problem tends not to occur in high-dimensional data spaces where this kind of granularity is less likely. However, then it gives rise to the problem of neighbouring anomalous clusters ( as illustrated in Figure 6 (c-II, d-II) ), which individually appear to be typical, or of limited suspicion (due to the presence of other neighbouring anomalous clusters), yet, their co-occurrence is highly anomalous.

Figure 6 (f-II) provides another situation in which false negatives increase because of the clustering step. This bivariate dataset contains 1,001 data points. The data points gathered at the leftmost upper corner represent a typical class covering 1,000 data points, and the isolated data point at the rightmost bottom corner represents an anomaly. Since this typical class of 1,000 data points is closely compacted, it gives rise to only 14 clusters through the Leader algorithm. Altogether, the dataset forms 15 clusters with the one created by the isolated point located at the rightmost bottom corner. Even though the original dataset contains 1,001 data points, the algorithm considers only 15 data points (a representative member from each cluster) for calculating the anomalous threshold. Now, this number is not large enough to yield a stable estimate for the anomalous threshold. Due to this ignorance of the density of the original dataset, it now fails to detect the obvious anomalous point at the leftmost bottom corner.

## 3.3 Problem with Threshold Calculation

A companion R package (Fraley 2018) is available for the algorithm proposed by Wilkinson (2017). According to the R package implementation, the current version of the HDoutliers algorithm uses the next potential candidate for anomalies in calculating the anomalous threshold, in each iteration of the bottom-up searching algorithm. This approach causes an increase in the false detection rate under certain circumstances. We avoid this limitation in our proposed algorithm.

# 4 Proposed Improved Algorithm: `stray` Algorithm

In this section, we propose an improved algorithm for anomaly detection in high dimensional data. Our proposed algorithm is intended to overcome the limitations of the HDoutliers algorithm and thereby enhance its capabilities.

## 4.1 Input to the `stray` Algorithm

An input to the stray algorithm is a collection of data instances where each data instance can be a realisation of only one attribute or a set of attributes (also referred to using terms such as features, measurements and dimensions). In this study, we limit our discussion to quantitative data; therefore, an input can be a vector, matrix or data frame of $d(\geq 1)$ numerical variables, where each column corresponds to an attribute and each row corresponds to an observation of these attributes. The focus is then to detect anomalous instances (rows) in the dataset.

## 4.2 Normalise the Columns

Since the stray algorithm is based on the distance definition of an anomaly, nearest neighbour distances between data instances in the high-dimensional data space are the key information for the algorithm to detect anomalies. However, variables with large variance can exert disproportional influence on Euclidean distance calculations (Wilkinson 2017). To make the variables of equivalent weight, the columns of the `data` are first normalised such that the data are bounded by the unit hypercube. This normalisation is commonly referred as *min-max normalisation*, which involves a linear transformation of the original data, with the result data ranging from 0 to 1. This type of transformation does not change the distribution or squeeze points together masking anomalies.

## 4.3 Nearest Neighbour Searching

In the `stray` algorithm, after the columns of the dataset are normalised, it calculates the k-nearest neighbour distance with the maximum gap for each and every instance. By using this measure, we were able to address the aforementioned limitations of the HDoutliers algorithm.

For each individual observation, the algorithm first calculates the *k*-nearest neighbour distances, $d_{i,KNN}$, where $i = 1, 2, ..., k$. Then, it calculates the successive differences between distances, $\Delta_{i,KNN}$. Next, it selects the k-nearest neighbour distance with the maximum gap, $\Delta_{i,max}$. Figure 3 illustrates how these steps help our improved algorithm to detect anomalous points or anomalous clusters of points.

In Figure 3 (a), the dataset contains only one anomaly at (15, 16.5). For this dataset, the nearest neighbour distance can differentiate the anomalous point from the remaining typical points because the nearest neighbour distance for the anomalous point is significantly larger (14.8) than that for the remaining typical points. Figure 3 (b) shows the change in the k-nearest neighbour distances of the anomaly at (15, 16.5). For this dataset, the k-nearest neighbour distance with the maximum gap occurs when $k = 1$. The second dataset, in Figure 3 b), has three anomalies around (15, 16.5). If we calculate only the nearest neighbour distances for each observation, then the three anomalies are not distinguishable from the typical points since their values are very small (0.7) compared with that of most typical points with nearest neighbour distances at around (0.0015 to 2.5). However, the three anomalies are distinguishable from their typical points with respect to the k-nearest neighbour distances with the maximum gap (Figure 3 (d)). For the three anomalies in Figure 3 (d), the third nearest neighbour distance has the maximum gap (Figure 3 e)) and the three points are now easily distinguished as anomalies, with respect to k-nearest neighbour distances with the maximum gap. Therefore, by using k-nearest neighbour distances with the maximum gap, the stray algorithm gains the ability to detect both anomalous singletons and micro clusters. Through this approach, we are able to reduce the false detection rate and thereby address the limitations of the HDoutliers algorithm, while gaining the ability to detect micro clusters. This is also a very simple, but clever, investment as compared with the time taken by the leader algorithm to form small clusters to detect micro clusters (especially for datasets with large dimensions), in the HDoutliers algorithm. Further, for each point, the corresponding k-nearest neighbour distances with the maximum gap act as an anomalous score to indicate the degree of being an anomaly.

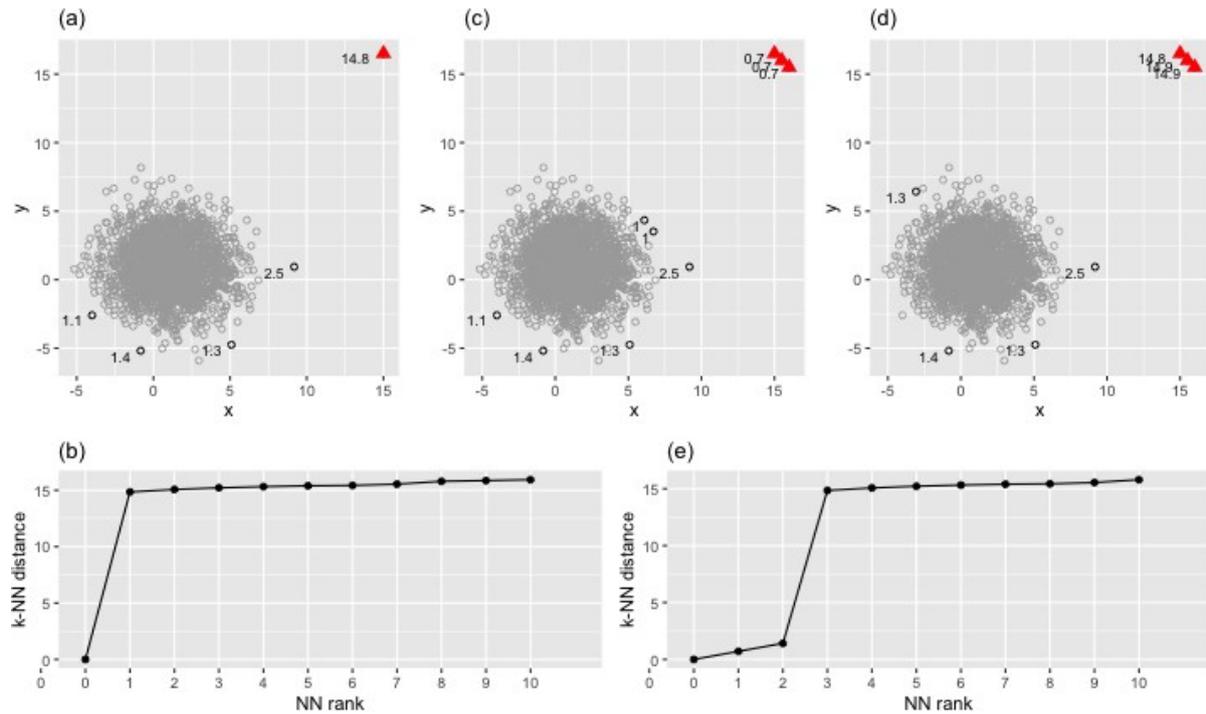

**Figure 3:** *Difference between the nearest neighbour distance and the k-nearest neighbour distance with the maximum gap. (a) Dataset contains only one anomaly at (15, 16.5). Nearest neighbour distances are marked. (b) Change in the k-nearest neighbour distances of the anomaly. (c) Dataset contains micro cluster around (15, 16.5). Nearest neighbour distances are marked. (d) Dataset contains micro cluster around (15, 16.5). For the three anomalies, the third nearest neighbour distance has the maximum gap. (e) Change in the k-nearest neighbour distances of an anomaly from micro cluster around (15, 16.5). Anomalies are represented by red triangles and black dots correspond to the typical behaviour.*

In the current study, we consider both exact and approximate k-nearest neighbour searching techniques. Brute force search involves going through every possible paring of points to detect k-nearest neighbours for each data instance, and therefore, exact k-nearest neighbours are explored. Conversely, k-dimensional trees (k-d trees) employ spatial data structures that partition space to allow efficient access to a specified query point (Elseberg et al. 2012b). Therefore, it involves searching approximate k-nearest neighbours around a specified query point.

In the current algorithm, parameter *k*, which determines the size of the neighbourhood, is introduced as a user-defined parameter that can be selected according to the application. One way to interpret the role of *k* in the stray algorithm is to view it as the minimum possible size for a typical cluster in a given dataset. If the size of an anomalous cluster is less than *k*, it will be detected as a micro cluster by the stray algorithm. The choice of *k* has different effects across different dimensions and sizes of data (Campos et al. 2016). We can set *k* to 1 if no micro clusters are present in the dataset and thereby focus on local and global anomalous points. High *k* values are recommended for datasets with high dimensions because of the curse of dimensionality.

### 4.4 Threshold Calculation

Anomalous scores assign each point a degree of being an anomaly. However, for certain applications it is also important to categorise typical and anomalous points for the subsequent root-cause analysis. Ideally, we prefer a universal threshold to unambiguously distinguish anomalous points from typical points. Following Schwarz (2008), the HDoutliers algorithm (Wilkinson 2017) defines an anomalous threshold based on extreme value theory, a branch of probability theory that relates to the behaviour of extreme order statistics in a given sample (Galambos, Lechner & Simiu 2013).

The anomalous threshold calculation in Schwarz (2008); Burridge & Taylor (2006) and Wilkinson (2017) is an application of Weissman's spacing theorem (Weissman 1978) (Theorem 4.1) that is applicable to the distribution of data covered by the maximum domain of attraction of a Gumbel distribution. This requirement is satisfied by a wide range of distributions, ranging from those with light tails to moderately heavy tails that decrease to zero faster than any power function (Embrechts, Klüppelberg & Mikosch 2013). Examples include the exponential, gamma, normal and log-normal distributions with exponentially decaying tails.

Let $X_1, X_2, \ldots, X_n$ be a sample from a distribution function $F$ and let $X_{1:n} \leq X_{2:n} \leq \cdots \leq X_{n:n}$ be the order statistics. The available data are $X_{1:n}, \ldots, X_{k:n}$ for some fixed $k$.

**Theorem 4.1** (Spacing Theorem). *(Proposition 1 in Burridge & Taylor (2006), p.6 and Theorem 3 in Weissman (1978), p.813; the notations have been changed for consistency in this paper)*

*Let $D_{i,n} = X_{i:n} - X_{i+1:n}$, $(i = 1, \ldots, k)$ be the spacing between successive order statistics. If $F$ is in the maximum domain of attraction of the Gumbel distribution, the spacings $D_{i,n}$ are asymptotically independent and exponentially distributed with mean proportional to $i^{-1}$.*

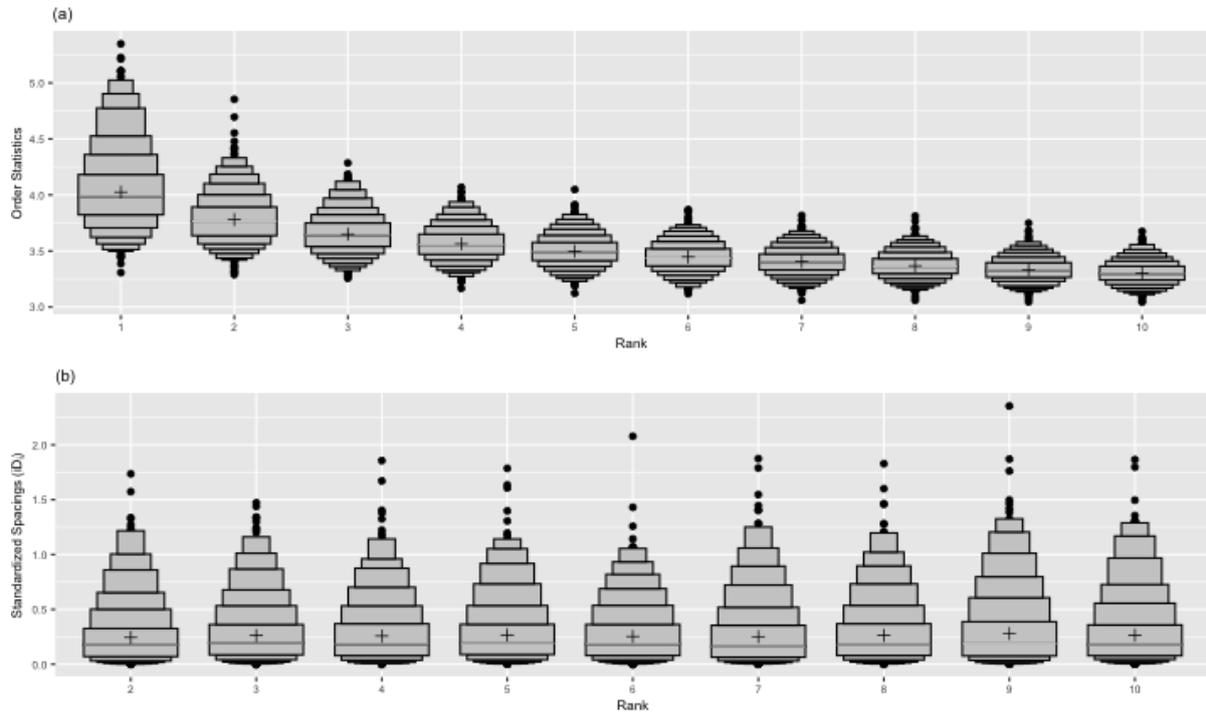

**Figure 4:** *(a) Distribution of the descending order statistics $X_{i:n}$ and (b) distribution of the standardised spacings $iD_{i,n}$ for $i \in \{1, \ldots, 10\}$ for $1,000$ samples each containing $20,000$ random numbers from the standard normal distribution.*

We illustrate this theorem using Figure 4, which shows the distribution of the descending order statistics ($X_{i:n}$) and the standardized spacings, ($iD_{i,n}$), for $i \in \{1, \ldots, 10\}$ for $1,000$ samples each containing $20,000$ random numbers from the standard normal distribution. Figure 4 (a) shows the distribution of $X_{i:n}$ with means of $X_{i:n}$ depicted as black crosses. The gaps between consecutive black crosses give the spacings between higher-order statistics ($D_{i,n}$). We note that the normal distribution is in the maximum domain of attraction of the Gumbel distribution and that this example contains no outliers. A consequence of Theorem 4.1 is that the standardised spacings ($iD_{i,n}$) for ($i = 1, \ldots, K$), are approximately iid (Burridge & Taylor 2006). Figure 4 (b) shows the distribution of the standardised spacings ($iD_{i,n}$) for ($i = 1, 2, ..., 10$) for $1,000$ samples of size $20,000$. Each letter-value box plot (Hofmann, Wickham & Kafadar 2017) exhibits approximately the shape of an exponential distribution.

Following Schwarz (2008), Burridge & Taylor (2006) and Wilkinson (2017), we start our anomalous threshold calculation from a subset of the points covering 50 per cent of them with the smallest anomalous scores under the assumption that this subset contains the anomalous scores corresponding to typical data points and the remaining subset contains the scores corresponding to the possible candidates for anomalies. Following the Weissman spacing theorem, it then fits an exponential distribution to the upper tail of the outlier scores of the first subset, and then computes the upper $1 - \alpha$ points of the fitted cumulative distribution function, thereby defining an anomalous threshold for the next anomalous score. Then, from the remaining subset it selects the point with the smallest anomalous score. If this anomalous score exceeds the cut-off point, it flags all the points in the remaining subset as anomalies and stops searching for anomalies. Otherwise, it declares the point as a typical point and adds it to the subset of the typical points. It then updates the cut-off point, including the latest addition. This searching algorithm continues until it finds an anomalous score that exceeds the latest cut-off point. This algorithm is known as a 'bottom-up searching' algorithm in Schwarz (2008). This threshold calculation is performed under the assumption that the distribution of k-nearest neighbours with the maximum gap is in the maximum domain of attraction of the Gumbel distribution, which covers a wide range of distributions.

### 4.5 Output

In `stray`, anomalies are measured in two scales: (1) binary classification and (2) outlier score. Under binary classification, data instances are classified either as typical or anomalous using the data-driven anomalous threshold based on the extreme value theory. This type of classification is important if the subsequent steps of the data analysis process are automated. The stray algorithm also assigns an anomalous score to each data instance to indicate the degree of outlierness of each measurement. These anomalous scores allow the user to rank and select the most serious or relevant anomalous points for root-cause analysis and taking immediate precautions. The HDoutliers algorithm (Wilkinson 2017), which provides only a binary classification, does not directly allow the user to make such a choice to direct their attention to more significant anomalous instances. Conversely, various methods proposed in the literature provide anomalous scores, but the anomalous threshold is user defined and application specific (Madsen 2018). The output produced by `stray` is an all-in-one solution encapsulating necessary measurements of anomalies for further actions.

## 5 Experiments

The HDoutliers algorithm is a powerful algorithm in the current state-of-the-art methods for detecting anomalies in high-dimensional data. The focus of the stray algorithm is to address some of the limitations of the HDoutliers algorithm that hinder its performance under certain circumstances. Here, we perform an experimental evaluation on the accuracy and computational efficiency of our stray algorithm relative to the HDoutliers algorithm. While these examples are fairly limited in number and are mostly limited to bivariate datasets, they should be viewed only as simple illustrations of the key features of the stray algorithm that outperforms the HDoutliers algorithm.

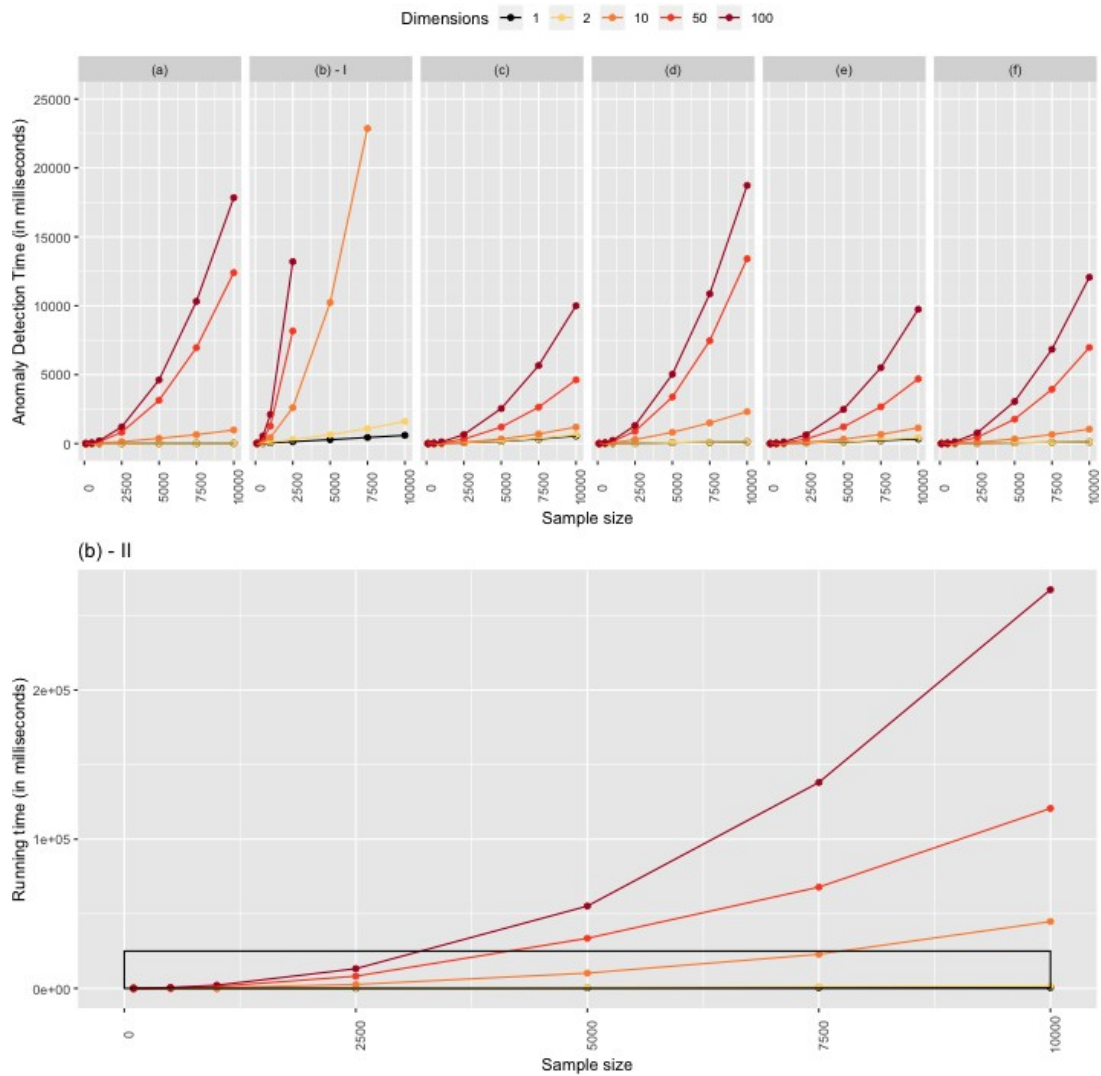

**Figure 5:** *Scalability Performance. (a) HDoutliers algorithm without clustering step, (b-I) HDoutliers algorithm with clustering step, (c) stray algorithm with brute force nearest neighbour search using FNN R package implementation, (d) stray algorithm with kd-trees nearest neighbour search using 'FNN' R package implementation, (e) stray algorithm with brute force nearest neighbour search using 'nabor' R package implementation, (f) stray algorithm with kd-trees nearest neighbour search using 'nabor' R package implementation. For clear comparison, only a part of the measurements of the full experiment is displayed in (b-I). (b-II) presents the full version of (b-I). Black frame in (b-II) covers the plotting region of (b-I).*

The first experiment (Figure 5) was designed to test the effect of the dimension, size of the data and the k-nearest neighbour searching method on running times of the different versions of the two algorithms: stray and HDoutliers.

The HDoutliers algorithm has two versions. The first version calculates nearest neighbour distance for each data instance and does not involve any clustering step prior to the nearest neighbour distance calculation. This version of the algorithm (version 1 of the HDoutliers, hereafter) is recommended for small samples (n<10,000). The second version uses the Leader algorithm to form several clusters of points and then selects a representative member from each cluster. The nearest neighbour distances are then calculated only for the selected representative members. Compared with version 1 of the HDoutliers algorithm (Figure 5 (a)), version 2 with the clustering step is extremely slow for higher dimensions (>10), and the running time increases more rapidly with increasing sample size. For clear comparison between the different versions of the two algorithms (stray and HDoutliers), only a part of the measurements of the full experiment of the second version of the HDoutliers algorithm is displayed in Figure 5 (b-I)). Figure 5 (b-II) presents the full version of Figure 5 (b-I). The additional clustering step in the second version of the HDoutliers algorithm, which is essential for detecting micro clusters, is extremely time-consuming, particularly with large samples with higher dimensions. Figure 5 (c)–(f) corresponds to the stray algorithm. In this experiment, to ascertain the influence from the k-nearest neighbour searching methods, we considered both exact (brute force) and approximate (kd-trees) nearest neighbour searching algorithms.

Many implementations of *k*-nearest neighbour searching algorithms are available for the R software environment. We considered FNN (Beygelzimer et al. (2019), Figure 5 (c) & (d)) and nabor (Elseberg et al. (2012a); Figure 5 (e) & (f)) R packages for our comparative analysis. R package nabor, wraps a fast k-nearest neighbour library written in templated C++. We noticed that searching $k-$ (> 1) nearest neighbours (Figure 5 (a), in this example *k* is set to 10) instead of only one ($k = 1$) nearest neighbour (Figure 5 (d)) increases the running time only slightly as the number of instances is increased. The results in both Figure 5 (a) and Figure 5 (d) are based on approximate nearest neighbour distances using the kd-trees nearest neighbour searching algorithm. We observed that the kd-trees implementation in nabor package (Figure 5 (f)) is much faster than the FFN package implementation (Figure 5 (d)). Surprisingly, as the dimension increases, the running time of the stray algorithm with kd-trees (Figure 5 (d), (f)) increases much more quickly than that of the brute force algorithm, which involves searching every possible pairing of points to detect *k*-nearest neighbours for each data instance (Figure 5 (c), (e)). Other studies (Kanungo et al. 2002) have also reported a similar result for many algorithms based on kd-trees and many variants. This could be due to the parallelisability and memory access patterns of the two searching mechanisms. The brute force algorithm is easily parallelisable because it involves independent searching of all possible candidates for each data instance. In contrast, the kd-tree searching algorithm is naturally serial and therefore difficult to implement on parallel systems with appreciable speedup (Zhang 2017).

**Table 1:** *Performance metrics – False positive rates. The values given are based on 100 iterations and the mean values are reported. Different versions of the two algorithms (stray and Hdoutliers) are applied on datasets where each column is randomly generated from the standardised normal distribution. All the datasets are free from anomalies HDoutliers WoC: HDoutliers algorithm without clustering step; HDoutliers WC: HDoutliers algorithm with clustering step.*

| Method | dim | 100 | 500 | 1000 | 2500 | 5000 | 7500 | 10000 |
| --- | --- | --- | --- | --- | --- | --- | --- | --- |
| HDoutliers WoC | 1 | 0.017 | 0.011 | 0.008 | 0.007 | 0.005 | 0.005 | 0.004 |
| HDoutliers WoC | 10 | 0.002 | 0.002 | 0.002 | 0.002 | 0.002 | 0.002 | 0.002 |
| HDoutliers WoC | 100 | 0.001 | 0.001 | 0.001 | 0.001 | 0.001 | 0.001 | 0.001 |
| HDoutliers WC | 1 | 0.036 | 0.024 | 0.024 | 0.019 | 0.017 | 0.014 | 0.013 |
| HDoutliers WC | 10 | 0.006 | 0.006 | 0.006 | 0.005 | 0.005 | 0.005 | 0.005 |
| HDoutliers WC | 100 | 0.003 | 0.003 | 0.003 | 0.003 | 0.003 | 0.003 | 0.003 |
| stray - brute force | 1 | 0.006 | 0.003 | 0.002 | 0.002 | 0.002 | 0.001 | 0.001 |
| stray - brute force | 10 | 0.001 | 0.001 | 0.001 | 0.001 | 0.001 | 0.001 | 0.000 |
| stray - brute force | 100 | 0.000 | 0.000 | 0.000 | 0.000 | 0.000 | 0.000 | 0.000 |
| stray - FNN kd-tree | 1 | 0.006 | 0.003 | 0.002 | 0.002 | 0.002 | 0.001 | 0.001 |
| stray - FNN kd-tree | 10 | 0.001 | 0.001 | 0.001 | 0.001 | 0.001 | 0.001 | 0.000 |
| stray - FNN kd-tree | 100 | 0.000 | 0.000 | 0.000 | 0.000 | 0.000 | 0.000 | 0.000 |
| stray - nabor brute | 1 | 0.006 | 0.003 | 0.002 | 0.002 | 0.002 | 0.001 | 0.001 |
| stray - nabor brute | 10 | 0.001 | 0.001 | 0.001 | 0.001 | 0.001 | 0.001 | 0.000 |
| stray - nabor brute | 100 | 0.000 | 0.000 | 0.000 | 0.000 | 0.000 | 0.000 | 0.000 |
| stray - nabor kd-tree | 1 | 0.006 | 0.003 | 0.002 | 0.002 | 0.002 | 0.001 | 0.001 |
| stray - nabor kd-tree | 10 | 0.001 | 0.001 | 0.001 | 0.001 | 0.001 | 0.001 | 0.000 |
| stray - nabor kd-tree | 100 | 0.000 | 0.000 | 0.000 | 0.000 | 0.000 | 0.000 | 0.000 |

Following (Wilkinson 2017), we evaluated the false positive rate (typical points incorrectly identified as anomalies) of the stray algorithm by running it many times on random data. The values presented in Table 1 are based on 1000 iterations and the mean values are reported. Different versions of the two algorithms (stray and Hdoutliers) were applied on datasets where each column is randomly generated from the standardised normal distribution. In each test, the critical value, $\alpha$, was set to 0.05. Compared with the HDoutliers algorithm, low false positive rates were achieved for the stray algorithm across all dimensions and sample sizes. Unlike in the HDoutliers algorithm (Unwin 2019), in stray a much smaller false detection rate was observed even for the small datasets with smaller dimensions. No difference was observed across different versions of the stray algorithm with different nearest neighbour searching mechanisms and their different implementations.

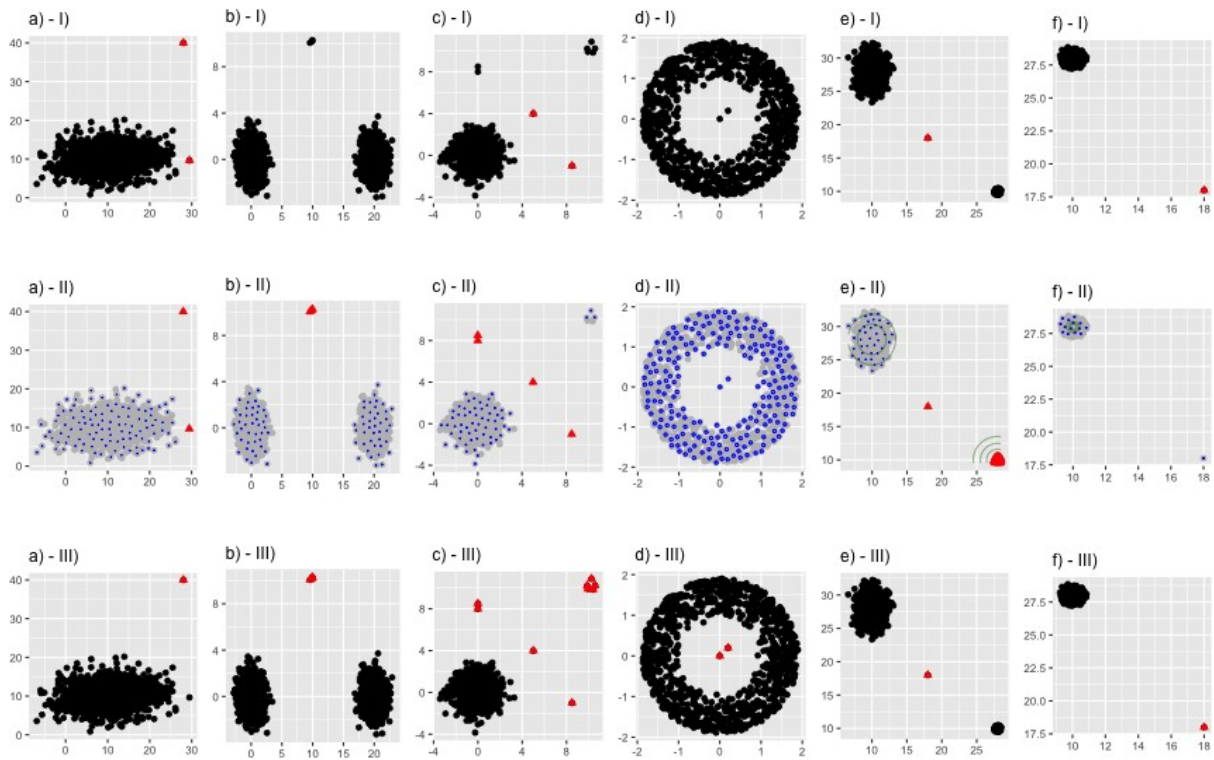

**Figure 6:** *Algorithm performance. (a) The top panel shows the results of the HDoutliers algorithm without a clustering step. (b) The middle panel shows the results of the HDoutliers algorithm with a clustering step. The representative member selected from each cluster formed by the Leader algorithm are marked in blue colour. (c) The bottom panel shows the results of the improved algorithm with brute force k-nearest neighbour searching. The detected anomalies are marked as red triangles.*

Figure 6 demonstrates how the stray algorithm outperforms the two versions of the HDoutliers algorithm under different circumstances. These limited set of examples were selected with the aim of highlighting some of the key feature of the stray algorithm:

(1) All three algorithms were able to correctly capture the anomalous point at the rightmost upper corner of Figure 6 (a)- I). However, the two versions of the HDoutliers algorithm tend to generate some false positives, particularly with the small dimensions.

(2) Figure 6 (b)- III) shows its ability to deal with multimodal typical classes. The two clusters at the bottom of the graph represent two typical classes. Only the second version of the HDoutliers algorithm (Figure 6 (b)- II) that utilises the clustering step was able to detect the top-centred micro cluster that contains three anomalous data instances. However, forming small clusters prior to the distance calculation is not always helpful in detecting micro clusters.

(3) Figure 6 (c)- II) shows a situation where even the second version of the HDoutliers algorithm fails in detecting micro clusters. The Leader algorithm in the HDoutliers algorithm uses a very small ball of a fixed radius to form clusters, and therefore, it now fails to capture the five points into a single cluster and instead generates three small clusters that are very close to one another. Both versions of the HDoutliers algorithm now fail to detect the micro cluster at the rightmost upper corner, because the dataset violates one of the major requirements of isolation of anomalous points or anomalous clusters. In stray, the value of *k* was set to 10. One can interpret the value of *k* as the maximum permissible size for a micro cluster. That is, for a small cluster to be a micro cluster, the number of data points in that cluster should be less than *k*. Otherwise, the cluster is considered a typical cluster.

(4) Figure 6 (d)- III) demonstrates the ability of detecting inliers. The HDoutliers algorithm also has this ability of detecting inliers only when there are isolated inliers that are free from anomalous neighbours. Both versions of the HDoutliers algorithm fail to detect the two inliers since they are very close to one another and thereby jointly project them as being anomalous.

(5) As explained in Section 3.2, Figure 6 (e)- II) shows how the clustering step of the second version of the HDoutliers algorithm can misguide the detection process and thereby increase the rate of false positives. The dense areas of the dataset are marked with density curves. Two typical clusters are visible, one at the leftmost upper corner and the other at the rightmost bottom corner. An inlier is also present in between the two typical classes. After forming cluster through the Leader algorithm, only one representative member is selected from each cluster for the nearest neighbour distance calculation. The selected member is now isolated and earns a very high anomalous score, leading the entire typical cluster at the rightmost bottom corner with 1,000 points to be identified as anomalous. In contrast, the stray algorithm is free from these problems because it does not involve any clustering step prior to the nearest neighbour distance calculation.

(6) As explained in Section 3.2, Figure 6 (f)- II) shows how the clustering step can increase the rate of false negatives. This dataset contains one typical class that is closely compacted in substance (the leftmost upper corner) and an obvious anomaly at the rightmost bottom corner. Since the typical class is a dense cluster, only a few data points are selected from the typical class for the nearest neighbour calculation. In this example, the clustering step substantially down-samples the original dataset, leading to a huge information loss in the representation of the original dataset. The blue dots in Figure 6 (f)- II) represent the selected members from each cluster for nearest neighbour calculations. Now, the reduced sample size is not enough for a proper calculation of the anomalous threshold based on extreme value theory.

# 6 Usage

We applied our stray algorithm to a dataset obtained from an automated pedestrian counting system with 43 sensors in the city of Melbourne, Australia (City of Melbourne 2019; Wang 2018), to identify unusual pedestrian activities within the municipality. Identification of such unusual, critical behaviours of pedestrians at different city locations at different times of the day is important because it is a direct indication of a city's economic conditions, the related activities and the safety and convenience of the pedestrian experience (City of Melbourne 2019). It also guides and informs decision-making and planning. This case study also illustrates how the stray algorithm can be used to deal with other data structures, such as temporal data and streaming data using feature engineering.

## 6.1 Handling Temporal Data

For clear illustration, we limit our study period to one month from 1 December to 31 December 2018. Figure 7 shows the pedestrian counts at 43 locations in the city of Melbourne at different times of the day. Each scatterplot follows a negatively skewed distribution. In general, weekdays display a bimodal distribution, while weekends follow a unimodal distribution. Now, the aim is to detect days with unusual behaviours. Since this involves a large collection of scatterplots, manual monitoring is time-consuming and unusual behaviours are difficult to locate by visual inspection.

Detecting anomalous scatterplots from a large collection of scatterplots requires some pre-processing. In particular, to apply the stray algorithm, we need to convert this original dataset, with a large collection of scatterplots, into a high dimensional dataset. A simpler approach is to use features that describe the different shapes and patterns of the scatterplots. Computing features that describe meaningful shapes and patterns in a given scatterplot is straightforward with scagnostics (scatterplot diagnostics) developed by Wilkinson, Anand & Grossman (2005). For the current study, we select five features: outlying, convex, skinny, stringy and monotonic (Dang & Wilkinson 2014; Wilkinson, Anand & Grossman 2005). We specifically select these features to address our use-case. Once we extract these five features from each scatterplot, we convert our original collection of scatterplots into a high-dimensional dataset with five dimensions and 31 data instances. Figure 8 provides feature-based representation of the original collection of scatterplots. Each point in this high-dimensional data space corresponds to a single scatterplot (or a day) in the original collection of scatterplots. In this high-dimensional space, the stray algorithm detects two anomalous points and they are marked in red colour in Figure 8. The corresponding scatterplots (or days) are marked in red colour in Figure 7. Visual inspection also confirms the anomalous behaviour of these two scatterplots. Both days, 1 December 2018 and 31 December 2018, display an unusual rise later in the day. One selected day, 31 December 2018, is an obvious anomaly since it is the New Year's Eve, and the associated fireworks at Southbank in the city of Melbourne attract many thousands of visitors. Further investigations regarding 1 December 2018, reveal that there was a musical concert at the Melbourne Cricket Ground from 8.00 pm and the unusual rise later in the day could be due to the concert participants.

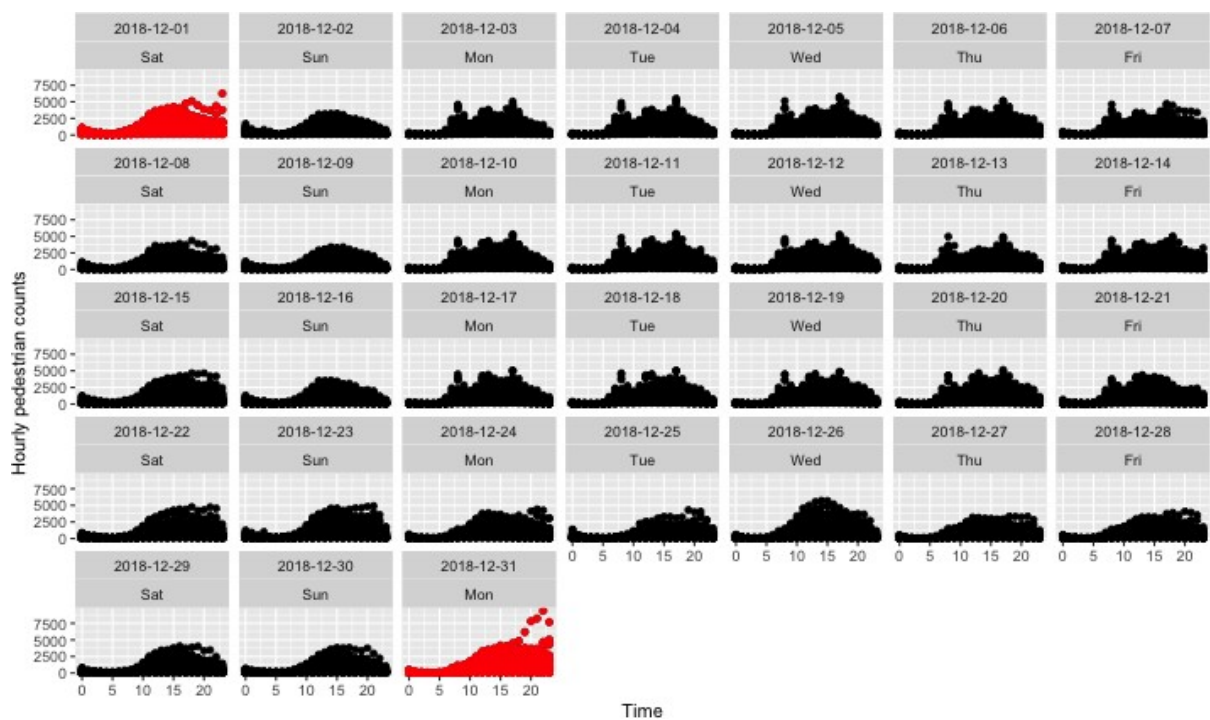

**Figure 7:** *Scatterplots of hourly pedestrian counts at 43 locations in the city Melbourne, Australia, from 1 December to 31 December 2018. Anomalous days detected by the stray algorithm using scagnostics are marked in red colour.*

After detecting the anomalous scatterplots or the days with anomalous pedestrian behaviours, further investigation is carried out for each day to detect the locations with anomalous behaviours within the selected day. Once we focus on one day, we obtain a collection of 43 time series with hourly pedestrian counts generated from the 43 sensors located at different geographical locations in the city (Figure 9). For this analysis, we extract seven time series features (similar to Talagala et al. 2019a; Hyndman, Wang & Laptev 2015) and convert the original collection of time series into a high dimensional data space with seven dimensions and 43 data instances (Figure 10). Now, each point in this high-dimensional space correspond to a single time series (or sensor) in Figure 9. The stray algorithm declares one point as an anomalous point in this high dimensional data space. This point corresponds to the sensor positions at Southbank in Melbourne where New Year's Eve fireworks attract millions of spectators annually.

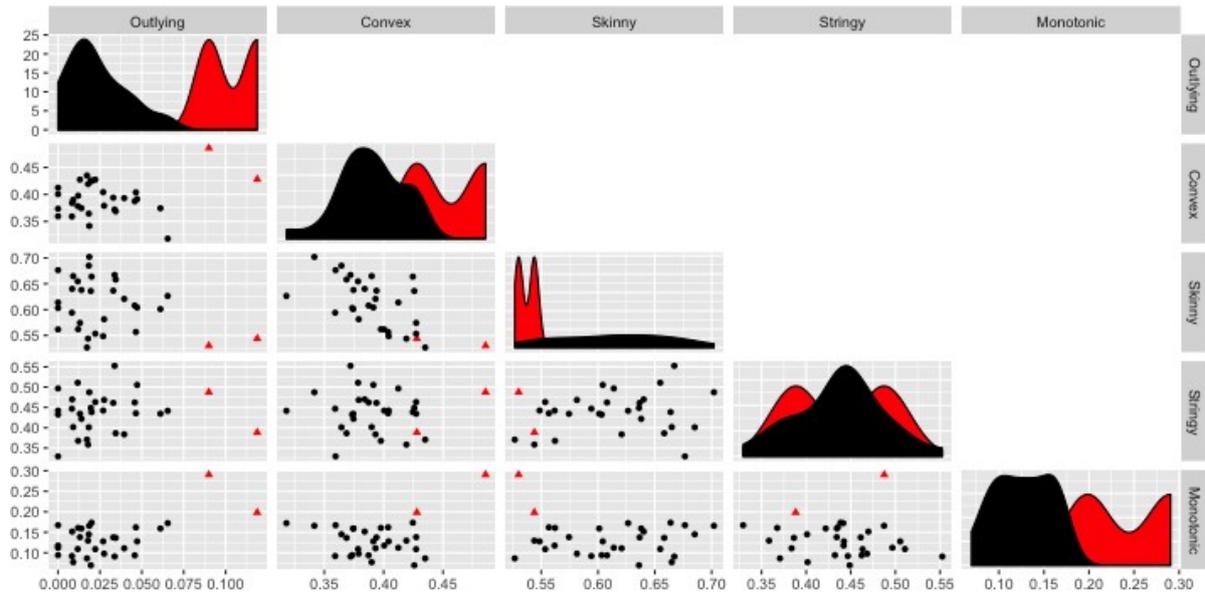

**Figure 8:** *Feature -based representation of the collection of scatterplots using Scagnostics. In each plot anomalies determined by the stray algorithm are represented by red colour.*

These types of findings play a critical role to make decisions about urban planning and management; to identify opportunities to improve city walkability and transport measures; to understand the impact of major events and other extreme conditions on pedestrian activity, and thereby assist in making decisions regarding security and resource requirements; and to plan and respond to emergency situations, etc.

### 6.2 Handling Streaming Data

Owing to the unsupervised nature of the stray algorithm, it can easily be extended for streaming data. A sliding window of fixed length can be used to deal with streaming data. Then, datasets in each window can be treated as a batch dataset (Talagala et al. 2019b) and the stray algorithm can be applied to each window to detect anomalies in the datasets defined by the corresponding window.

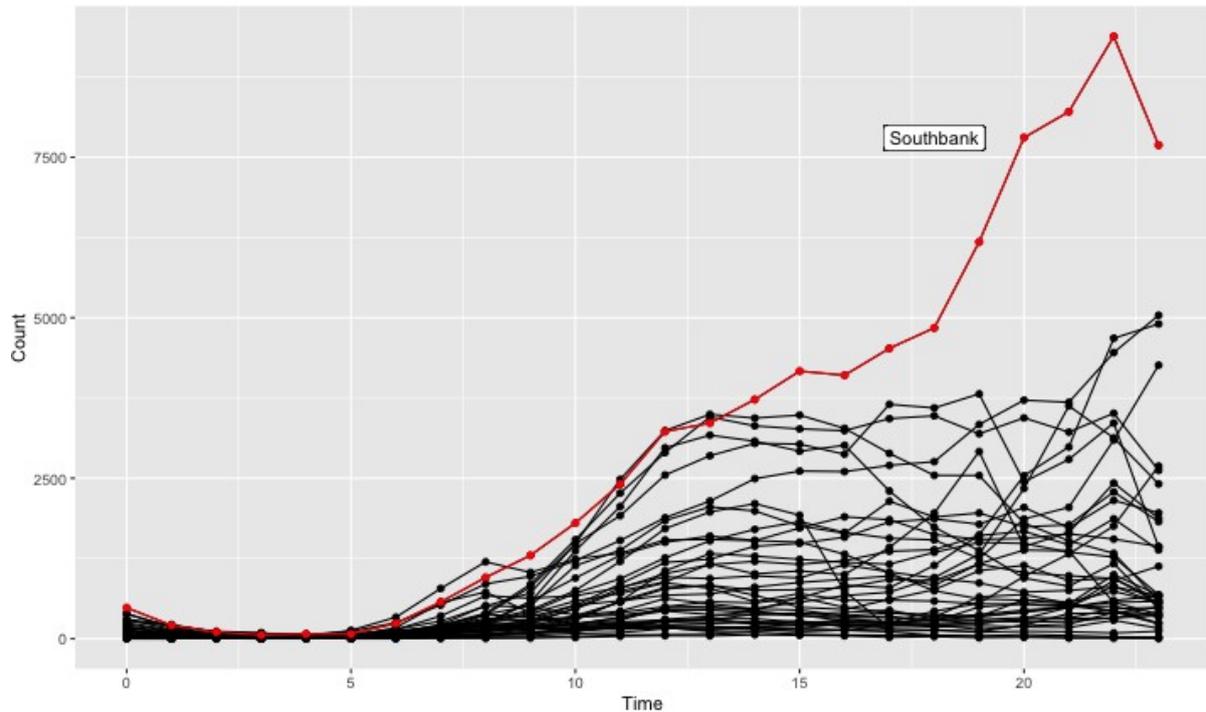

**Figure 9:** *Multivariate time series plot of hourly counts of pedestrians measured at 43 different sensors in the city of Melbourne, on 31 December 2018. The anomalous time series detected by the stray algorithm using time series features are marked in red colour.*

It also can be used to identify anomalous time series within a large collection of streaming temporal data. Let $W[t, t + w]$ represent a sliding window containing $n$ number of individual time series of length $w$. First, we extract $m$ features (similar to Hyndman, Wang & Laptev (2015) and Talagala et al. (2019b)) from each and every time series in this window. This step gives rise to an $n \times m$ feature matrix where each row now corresponds to a time series in the original collection of time series. Once we convert our original collection of time series into a high-dimensional dataset, we can apply the stray algorithm to identify anomalous points within this m-dimensional data space. The corresponding time series are then declared as anomalous series within the large collection of time series in the corresponding sliding window.

## 7 Conclusions and Further Research

The HDoutliers algorithm by Wilkinson (2017) is a powerful algorithm for detecting anomalies in high-dimensional data. However, it suffers from a few limitations that significantly hinder its ability to detect anomalies under certain situations. In this study, we propose an improved algorithm, the stray algorithm, that addresses these limitations. We define an anomaly here as an observation that deviates markedly from the majority with a large distance gap. We also demonstrate how the stray algorithm can assist in detecting anomalies present in other data structures using feature engineering. In addition to a label, the stray algorithm also assigns an anomalous score to each data instance to indicate the degree of outlierness of each measurement.

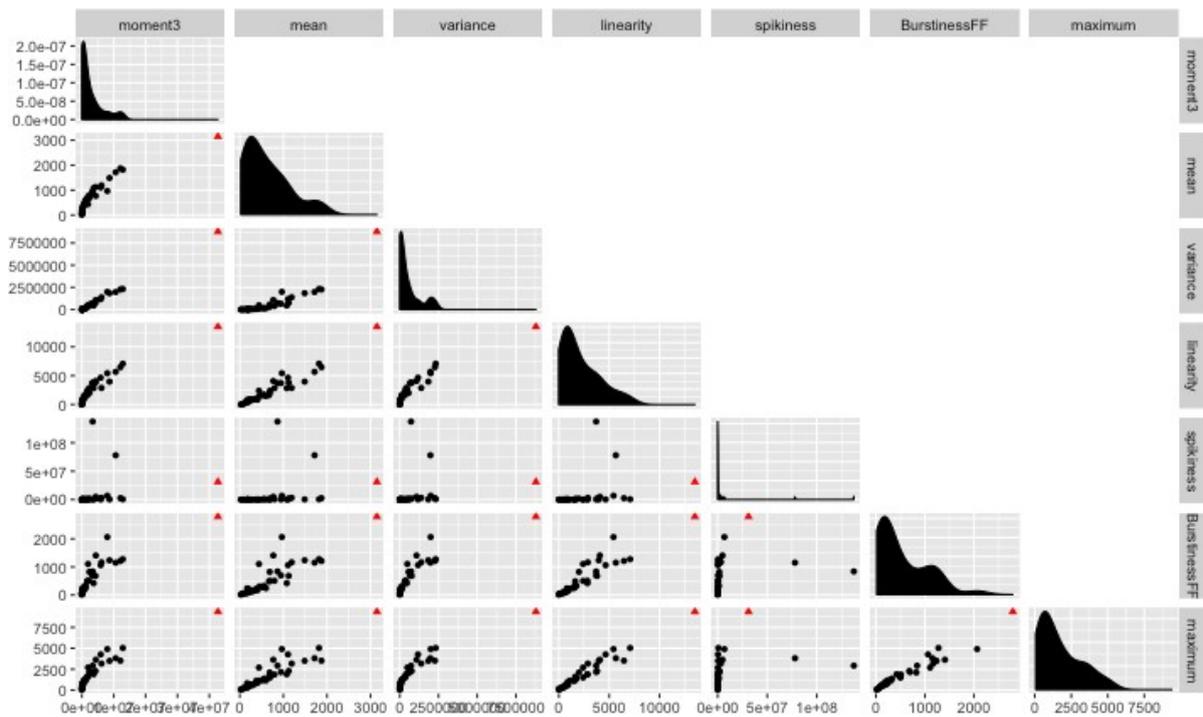

**Figure 10:** *Feature-based representation of the collection of time series on 31 December 2018. In each plot, anomalies determined by the stray algorithm are represented in red colour*

While the HDoutliers algorithm is powerful, we have provided several classes of counterexamples in this paper where the structural properties of the data did not enable HDoutliers to detect certain types of outliers. We demonstrated on these counterexamples that the stray algorithm outperforms HDoutliers, in terms of both accuracy and computational time. It is certainly common practice to evaluate the strength of an algorithm using collections of test problems with various challenging properties. However, we acknowledge that these counterexamples are not diverse and challenging enough to enable us to comment about the unique strengths and weaknesses of these two algorithms, nor to generalise our findings to conclude that stray is always the superior algorithm. This study should be viewed as an attempt to simulate further investigation on the HDoutliers algorithm and its successors, with the ultimate goal to achieve further improvements across the entire problem space defined by various high-dimensional datasets. An important open research problem is therefore to assess the effectiveness of these algorithms across the the broadest possible problem space defined by different datasets with diverse properties (Kang, Hyndman & Smith-Miles 2017). It is an interesting question to explore the impact of other classes of problems with various structural properties affect the performance of the stray algorithm and where its weaknesses might lie. This kind of instance space analysis (Smith-Miles et al. 2014) will enable further insights into improved algorithm design.

Anomaly detection problems commonly appear in many applications in different application domains. Therefore, it is hoped that different people with different knowledge levels will use the stray algorithm for many different purposes. Therefore, we expect future studies to develop interactive data visualisation tools that can enable exploring anomalies using a combination of graphical and numerical methods.